\documentclass{article}

\usepackage{arxiv}

\usepackage[utf8]{inputenc} % allow utf-8 input
\usepackage[T1]{fontenc}    % use 8-bit T1 fonts
\usepackage{hyperref}       % hyperlinks
\usepackage{url}            % simple URL typesetting
\usepackage{booktabs}       % professional-quality tables
\usepackage{amsfonts}       % blackboard math symbols
\usepackage{nicefrac}       % compact symbols for 1/2, etc.
\usepackage{microtype}      % microtypography
\usepackage{graphicx}
\graphicspath{{./}{./images/}}

\usepackage{amsmath,amssymb}
\usepackage{algorithmic}
\usepackage{xcolor}

\newcommand{\etal}{\textit{et al.}}

\title{A Conditional Generative Framework for Synthetic Data Augmentation in Segmenting Thin and Elongated Structures in Biological Images}

\author{
 Yi Liu \\
  Computer \& Information Science \\
  University of Delaware \\
  Newark, USA \\
  \texttt{yliu@udel.edu} \\
  \And
  Yichi Zhang \\
  Computer Science \\
  University of Virginia \\
  Charlottesville, USA \\
  \texttt{pcv9ha@virginia.edu} \\
}
\date{Nov 30, 2025}
\begin{document}
\maketitle

\begin{abstract}
Thin and elongated filamentous structures, such as microtubules and actin filaments, often play important roles in biological systems. Segmenting these filaments in biological images is a fundamental step for quantitative analysis. Recent advances in deep learning have significantly improved the performance of filament segmentation. However, there is a big challenge in acquiring high quality pixel-level annotated dataset for filamentous structures, as the dense distribution and geometric properties of filaments making manual annotation extremely laborious and time-consuming. To address the data shortage problem, we propose a conditional generative framework based on the Pix2Pix architecture to generate realistic filaments in microscopy images from binary masks. We also propose a filament-aware structural loss to improve the structure similarity when generating synthetic images. Our experiments have demonstrated the effectiveness of our approach and outperformed existing model trained without synthetic data. 
\end{abstract}

\section{Introduction}

Thin and elongated objects, such as microtubules and actin filaments, are ubiquitous in biological images, and these filamentous structures play important roles in biological systems. Segmenting these filamentous structures in biological images is a fundamental step to quantify their morphology, spatial organization, and underlying mechanisms. However, accurate segmentation remains challenging. Traditional image-processing methods are often unreliable because of high background noise and the complex geometry of filaments in microscopy images. Deep learning–based approaches have shown promising performance in image segmentation, but they require large, high-quality annotated datasets. Since filamentous structures are highly flexible and non-rigid, their shapes and configurations are virtually infinite. Therefore, robust segmentation of filamentous structures demands a significantly larger and more diverse dataset.

\begin{figure}[t]
\centerline{\includegraphics[width=1\linewidth]{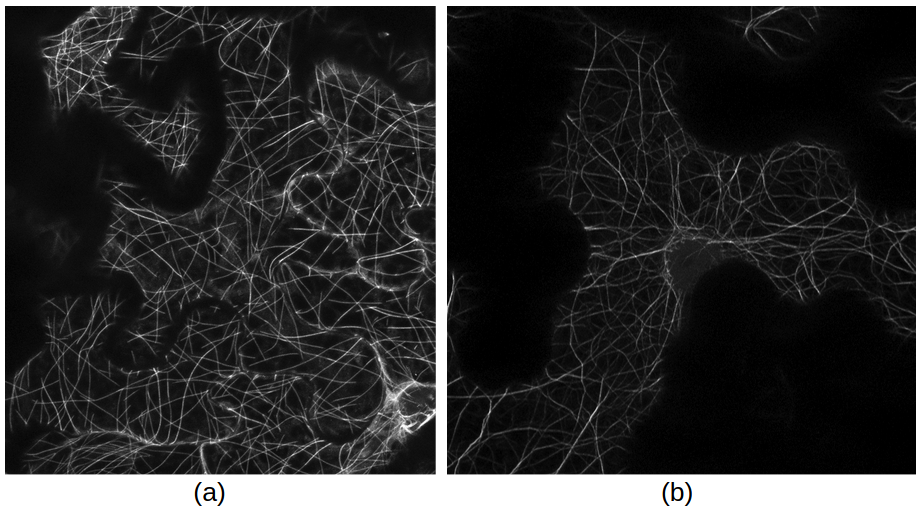}}
\caption{Examples of fluorescence microscopy images of filamentous structures. (a) Microtubules and (b) actin filaments exhibit distinct spatial organizations and densities. Each images contains hundreds or thousands of filaments, and these filaments are thin, elongated, and often overlap or intersect, making them difficult to trace accurately. 
Low signal-to-noise ratios and background fluorescence further obscure individual filaments.
Preparing accurate binary segmentation annotations for filaments is extremely laborious and time-consuming, requiring pixel-level manual tracing by experts.}
\label{actin_mt_example}
\end{figure}

In practice, obtaining large pixel-level annotated dataset for filamentous structure is extremely laborious and costly. As shown in Fig \ref{actin_mt_example}, each image can contain hundreds or thousands of filaments that need to be traced. The difficulty is further compounded by variations in imaging modalities, microscope settings, acquisition conditions and low signal-to-noise ratios, which make consistent annotation across datasets more challenging and requires significant manual effort from experts.

To address the challenges of acquiring annotated filament data, we propose a conditional generative framework to generate synthetic realistic filaments in microscopy images with paired binary mask as ground truth. Our model is built upon a conditional generative adversarial network\cite{isola2017image}, and it learns the mapping between binary filament masks and the corresponding microscopy images. To further enhance the model's ability to learn the unique structure of filaments, we introduce a filament-aware structural loss that enforces consistency along the filament skeletons. This improves model's ability to generate realistic connectivity and morphology of filaments. During inference time, the model will take binary masks of curves and generate realistic microscopic images of filaments. This will effectively expand the segmentation dataset and provide a cost-efficient alternative to manual annotation. Different from traditional augmentation methods such as rotation, flipping, perturbations, our generative framework learns the underlying distribution of microscopy appearance. It reproduces similar illumination, texture variations, and noise patterns similar to real fluorescence images while preserving filament geometry and continuity. 

We evaluated our proposed framework using microtubule and actin filament datasets and show that the inclusion of synthetic data significantly improves segmentation accuracy and generalization under limited-data conditions. The method achieves higher Intersection-over-Union (IoU) and skeleton IoU (SKIoU) compared to models trained solely on annotated data. Our qualitative results also demonstrate that models trained with synthetic augmentation produce smoother and more continuous filament predictions, particularly in noisy or low-contrast regions.

Our contributions are as follows:
\begin{itemize}
\item  We propose a conditional generative framework that generates realistic biological images of thin and elongated biological structures from binary masks, forming synthetic training dataset
\item  We introduce a filament-aware structural loss to preserve the unique properties of filament and improve the performance
\item  We conducted comprehensive experiments that demonstrate the effectiveness of generated synthetic data, outperforming exsiting methods.
\end{itemize}

\section{Related work}

\subsection{Filament Segmentation}

\begin{figure*}[t]
\centerline{\includegraphics[height=0.5\textheight, keepaspectratio]{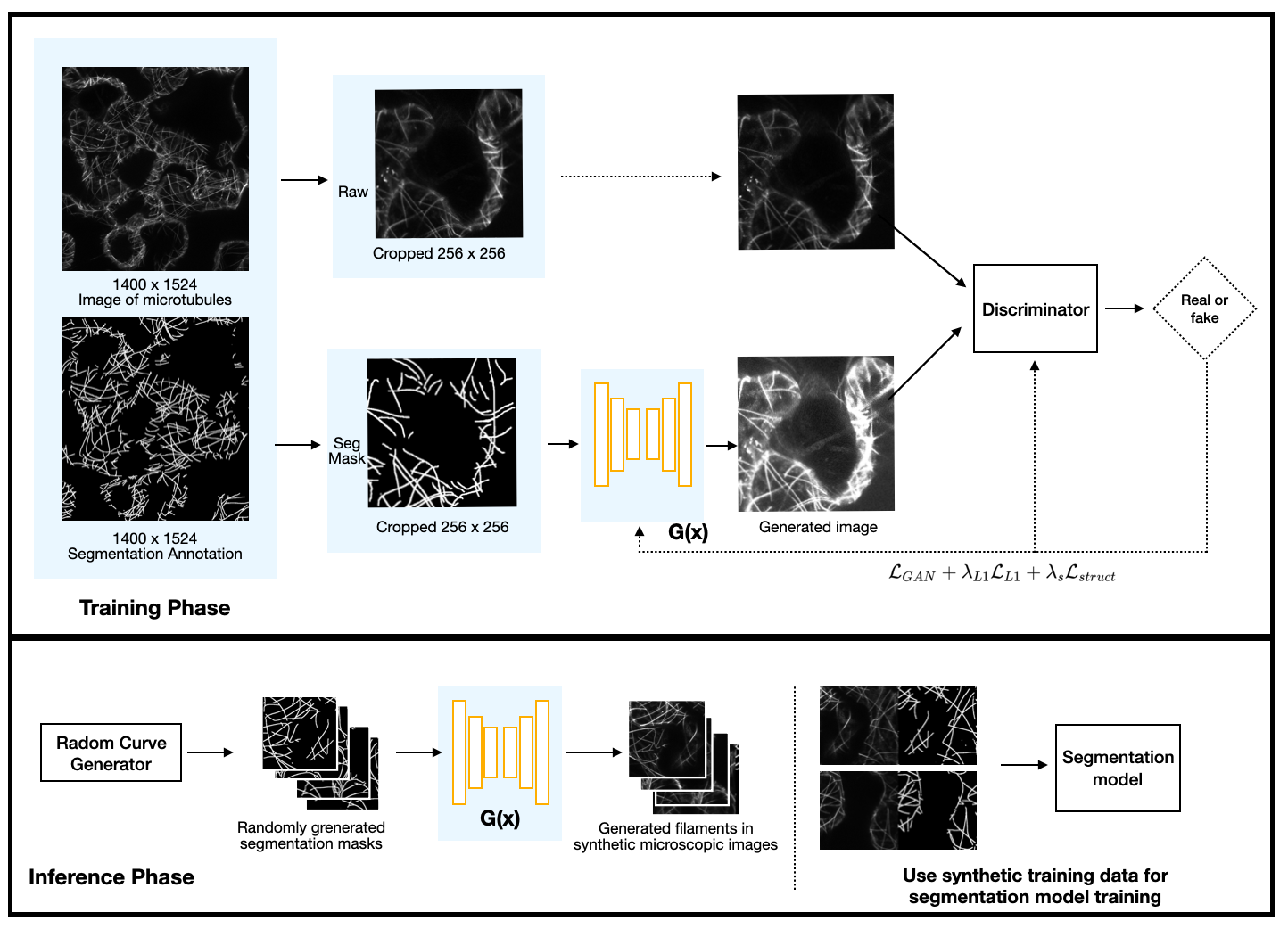}}
\caption{Overview of the proposed conditional generative framework for synthetic filament image generation and image segmentation.}
\label{fig}
\end{figure*}

Segmenting thin and elongated structures is fundamental in biomedical image analysis. Traditional image processing techniques have been widely used for segmenting filamentous structures in biological images~\cite{chang,paul,fuller2005filament,almi2015modified,soax,xu20143d,smith2010segmentation}. However, these approaches are highly sensitive to image quality, background noise, and filament density. They often require careful parameter tuning for each dataset. Recent advances in deep learning have substantially improved segmentation performance by learning hierarchical features from data. 
Deep learning–based segmentation models such as U-Net \cite{ronneberger2015u} and its variants \cite{liu2018densely, ye2025vsr, horiuchi2025deep, liu2019intersection, liu2022extraction} have been successfully applied to segment vessel-like and filamentous structures, achieving better results. Despite the strong performance, deep learning based models heavily rely on large, high-quality annotated datasets. However, creating segmentation datasets for filamentous structures is extremely labor-intensive and costly, as each image can contain hundreds of filaments that need to be traced manually, and may take hours to annotate one image \cite{liu2023pick, liu2018densely}. Moreover, due to the variations in imaging conditions and setups, segmentation models trained on limited dataset often failed to generalize across different imaging settings. Therefore, it is crucial to address the challenge of annotating filaments and enlarge the dataset, generalizing filament segmentation models to diverse imaging settings and biological contexts.

\subsection{Data Augmentation}
Imaging processing techniques such as rotation, flipping, and intensity adjustments have been widely used to augment data and improve segmentation performance \cite{chlap2021review, liu2018densely, liu2020quantifying}, but these methods only modify existing samples and fail to generate new filament shapes, complex novel patterns, or realistic background variations. 

Recent advances in deep generative models allow us to generate realistic synthetic images from learned data distributions. GAN-based \cite{goodfellow2020generative} augmentation has been widely explored to augment limited training data in biomedical image segmentation ~\cite{adjei2022examining, motamed2021data}. In \cite{thambawita2022singan}, Thambawita \etal proposed SinGAN-Seg to augment polyp images from a single training image. More recently, diffusion models have been a powerful alternative for generative augmentation. Nazir~\etal~\cite{nazir2025diffusion} introduced a diffusion-based augmentation framework for medical image segmentation that synthesizes anatomically consistent images conditioned on semantic masks.  
Similarly, the Polyp-DDPM model~\cite{dorjsembe2024polyp} proposed diffusion-based model to generate synthetic polyp textures and illumination variations for segmentation tasks.  

Despite the success of GAN- and diffusion-based data augmentation in various biomedical imaging tasks, most existing approaches focus on generating rigid objects or region-level structures such as organs or polyps, where shapes are relatively compact and well-defined. In contrast, filamentous structures are thin, elongated, and non-rigid, exhibiting high variance in orientation, length, curvature, and connectivity patterns. In this work, we specifically address the challenge of synthetic data generation for filamentous structures in microscopy images. Although diffusion-based generative models have become powerful and popular in recent work, they require substantially larger computational resources, longer training times, and much more data to converge effectively. While diffusion-based models have recently shown powerful performance in generative modeling, they require substantially more computational resources, longer training times, and larger datasets to converge effectively. In this work, we adopt a GAN-based conditional generative framework to provide a simpler, faster, and more resource-efficient solution tailored to filamentous microscopy images.

\section{Method}

From our discussion above, the challenge of constructing a large filamentous structure segmentation dataset is that filaments are non-rigid objects with variations in its lengths, curvatures and backgrounds. We propose a conditional generative model framework based to generate realistic microscopic style images of filaments from synthetic binary masks. As shown in figure \ref{fig}, our approach contains 3 stages: (1) Use limited annotated mask-image data to train the conditional generative model. (2) Generate randomized curves in binary images to represents mask, and use trained conditional generative model to generate realistic microscopic images. (3) training the segmentation network using both real and synthetic data.

\subsection{Conditional Generative Model}

We employed a conditional generative adversarial network (cGAN) \cite{isola2017image}  to model the translation between filament masks and microscopy images. The generator $G$ is used to produce realistic images $y_{\text{gen}}$ conditioned on input masks $x$, while the discriminator $D$ attempts to distinguish between real and generated image–mask pairs.  
The adversarial objective is defined as:
\begin{equation}
\mathcal{L}_{GAN}(G, D) = 
\mathbb{E}_{x, y}[\log D(x, y)] + 
\mathbb{E}_{x}[\log(1 - D(x, G(x)))].
\end{equation}

To generate real microscopy appearance, we also use $L_1$ reconstruction term to enforces pixel-level similarity between the synthesized image and the ground truth:
\begin{equation}
\mathcal{L}_{L1}(G) = \mathbb{E}_{x, y}[\|y - G(x)\|_1].
\end{equation}

\subsection{Filament-Aware Structural Loss}

To generate synthetic filaments, it is important to keep continuity of filaments. Since filaments are very thin, it is very sensitive to small shifts and variations. To enhance model's ability to learn the geometric similarity, we introduce filament-aware structural loss $\mathcal{L}_{struct}$.
We binarized both real and generated images, and skeletonized the images with morphological thinning. The loss is then defined as:
\begin{equation}
\mathcal{L}_{struct}(G) = \mathbb{E}_{x, y}[\|S(y) - S(G(x))\|_1],
\end{equation}
where $S(y)$ and $S(G(x))$ indicates the image after morphological thinning.

The following formula shows the final training objective with adversarial, reconstruction, and structure-aware terms:
\begin{equation}
\begin{split}
G^* = \arg \min_G \max_D \Big(
    \mathcal{L}_{GAN}(G, D)
    + \lambda_{L1} \mathcal{L}_{L1}(G) \\
    + \lambda_{s} \mathcal{L}_{struct}(G)
\Big),
\end{split}
\end{equation}
where $\lambda_{L1}$ and $\lambda_{s}$ is hyperparameter. 

\subsection{Synthetic Data Generation and Segmentation Training}

After training the cGAN on a small set of paired binary masks and real images, we first implemented a random filament mask generator to produce synthetic binary masks. Then we use trained cGAN to generate realistic microscopic images with synthetic binary mask, and creating augmentation dataset. For segmentation network setup, we employ a modified U-Net–based architecture \cite{liu2020quantifying} as the baseline model for its effectiveness in filament segmentation. 

\subsection{Implementation Details}

All experiments are implemented in PyTorch. The generator and discriminator follow the standard Pix2Pix architecture with 8 downsampling and upsampling blocks using LeakyReLU activations. All networks are trained using Adam optimizer with a learning rate of $1 \times 10^{-4}$, and batch size of 64. We set $\lambda_{L1}$ to 50 and $\lambda_{s}$ to 5. All images are cropped into $256 \times 256$ pixels for training. All models are trained on a single NVIDIA A100 GPU (40 GB).

\section{Experiments and Results}

\subsection{Datasets}

We evaluate the proposed framework on the cytoskeletal filament datasets in \cite{liu2018densely} which contains microtubules and actin filaments.
The dataset consists of fluorescence microscopy images captured under different imaging conditions and resolutions with 53 annotated microtubule images of size $1400 \times 1524$ and 10 annotated actin filament images of size $1400 \times 1524$.
We follow the same setup in \cite{liu2018densely}, use 25 microtubule images and 2 actin filament images in the datasets for training both cGAN and segmentation model, and testing on the rest 28 microtubule images and 8 actin images. 

\begin{figure}[t]
\centerline{\includegraphics[width=1\linewidth]{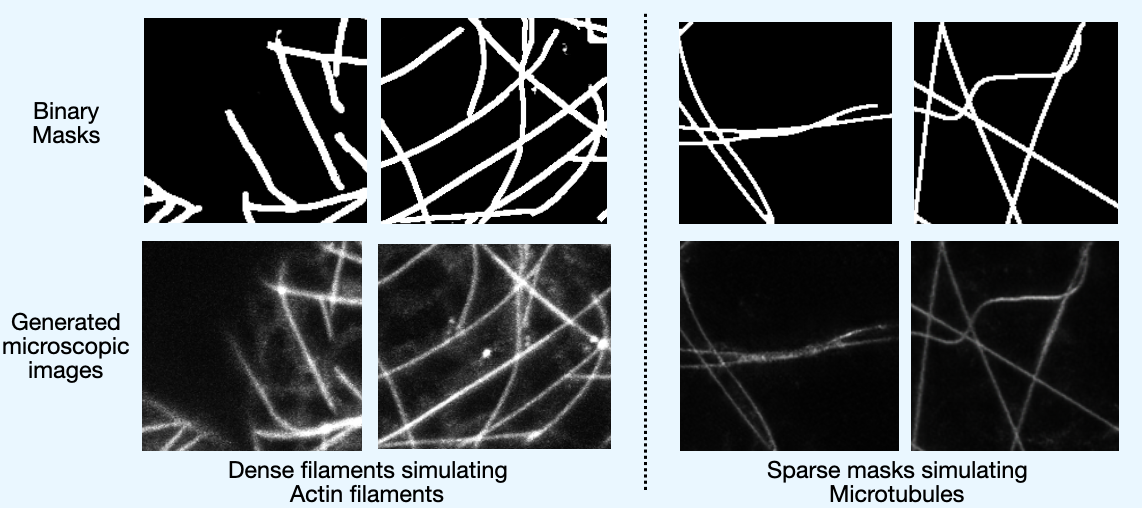}}
\caption{Examples of generated synthetic filament images. }
\label{sample_syn}
\end{figure}

Although the structure of actin filaments is more complex and denser than that of microtubules, the background variations are similar, and we generate denser filament mask to simulate actin filament structure. With trained cGAN, we generated 50000 $256 \times 256$ patches of microtubules and 50000 $256 \times 256$ patches of actin-like filaments. Fig. \ref{sample_syn} shows representative examples of the generated synthetic microscopy images.

\begin{figure}[t]
\centerline{\includegraphics[width=0.8\linewidth]{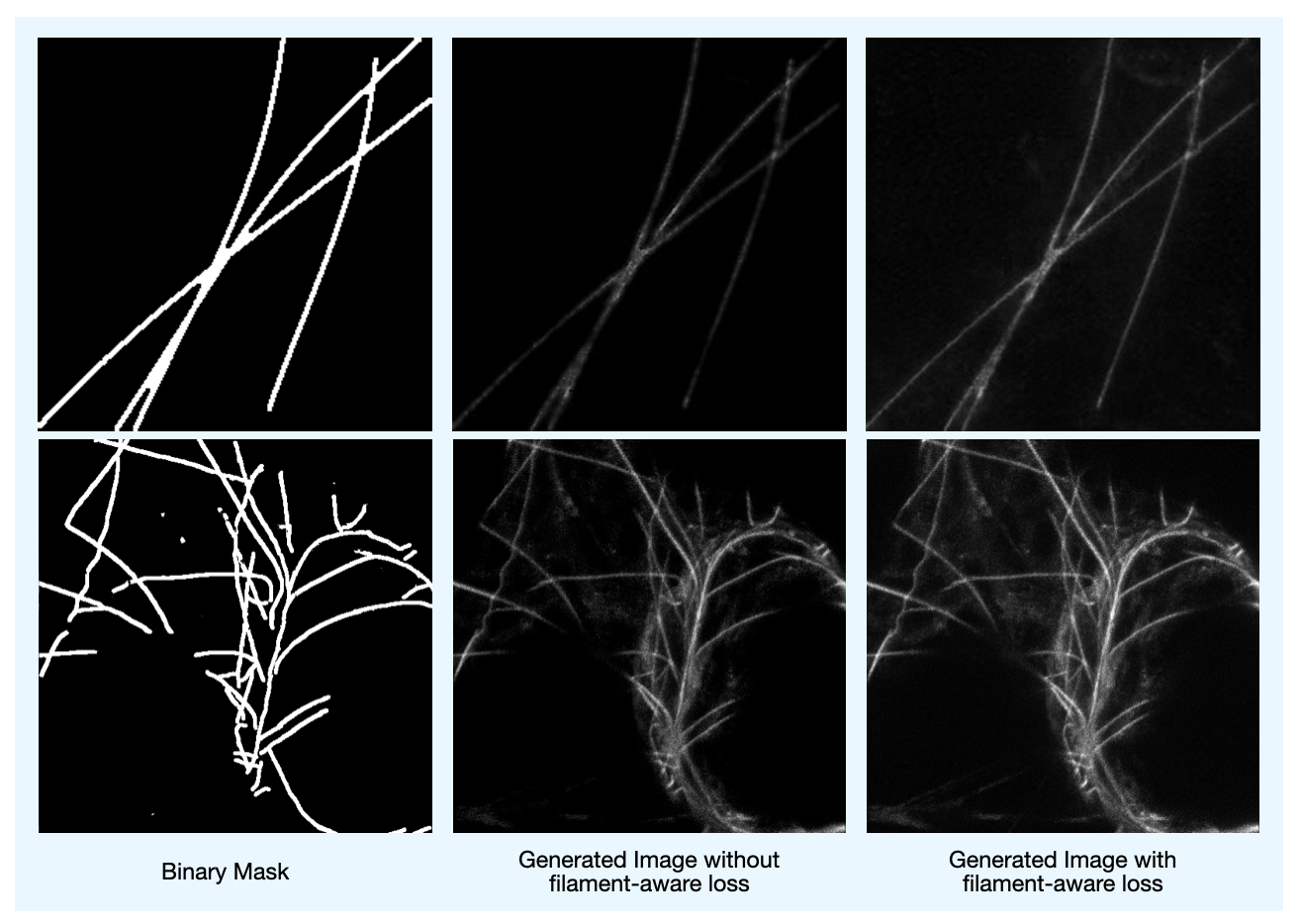}}
\caption{Examples of generated synthetic filament images. }
\label{compare}
\end{figure}

\begin{figure*}[t]
\centerline{\includegraphics[width=1\linewidth, keepaspectratio]{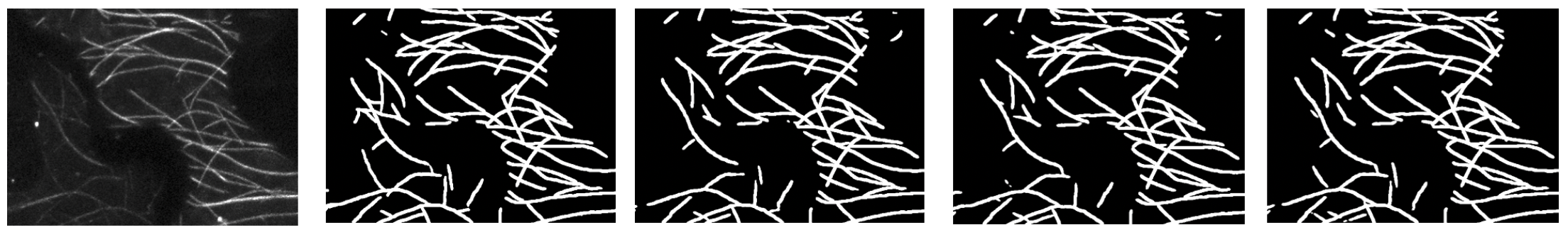}}
\caption{Segmentation of microtubules. From left to right: original image, ground truth, model trained with annotated images, annotated + synthetic dataset, annotated + synthetic with filament-aware loss}
\label{mt_quality}
\end{figure*}

\begin{figure*}[t]
\centerline{\includegraphics[width=1\linewidth, keepaspectratio]{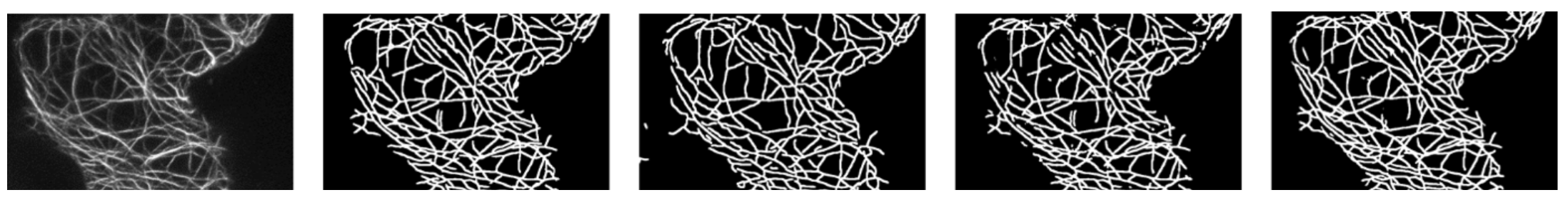}}
\caption{Segmentation of actin filaments. From left to right: original image, ground truth, model trained with annotated images, annotated + synthetic dataset, annotated + synthetic with filament-aware loss}
\label{actin_quality}
\end{figure*}

\subsection{Qualitative Results on Synthetic Filaments}

Fig. ~\ref{sample_syn} shows qualitative results of generated filaments. As we can see from the image, the model successfully generated realistic fluorescence microscopic images with filaments. It simulates the mechanism of fluorescence microscopy, where area with dense filaments have a  higher background intensity, and the brightness will be enhanced when there are multiple filaments overlapping each other. Fig. ~\ref{compare} shows the effectiveness of our proposed filament-aware loss. While both models can generate realistic filaments in microscopic images, the model with filament-aware loss generates less fragmented filaments. 

\subsection{Segmentation Results on Filament Dataset}

We adopt the Intersection-over-Union (IoU) and Skeleton IoU (SKIoU) metrics, following the definitions used in~\cite{liu2018densely}. IoU measures pixel-level overlap between predicted and ground-truth masks, while SKIoU focuses on structural similarity along filament skeletons.

We evaluate the effectiveness of our synthetic data generation framework on both microtubule and actin filament datasets. 
Table~\ref{tab:results} summarizes the quantitative results. We compare three experiments: (1) models trained on annotated datasets only, (2) models trained with our conditional generative model generated synthetic data, and (3) models trained with synthetic data generated with filament-aware structural loss. Traditional augmentation techniques such as image flipping, rotation are set as default setting. 

As we can see from table \ref{tab:results}, incorporating synthetic images significantly improves segmentation accuracy compared to training  on real annotated data.
On the microtubule dataset, adding synthetic data increases IoU from 0.9439 to 0.9527 and SKIoU from 0.9775 to 0.9849, indicating that the model benefits from enhanced structural diversity in training samples. We further incorporate samples generated with filament-aware loss model, and our method achieves the best performance with an IoU of 0.9542 and SKIoU of 0.9874.  
Similarly, on the actin filament dataset, our method yields consistent improvements, achieving an IoU of 0.9474 and SKIoU of 0.958.

Fig.~\ref{mt_quality} and Fig.~\ref{actin_quality} shows qualitative comparisons. Model trained synthetic data generate more continuous and topologically consistent filament predictions. Model trained with synthetic data generated with model with filament-aware loss shows best segmentation results with continuous filaments, especially in regions with clustered filaments and low signal areas.
These results demonstrate that our conditional generative framework effectively mitigates data scarcity and enhances segmentation robustness across different filament types.

\begin{table}[htbp]
\caption{Segmentation Results on Actin and Microtubule Datasets}
\begin{center}
\begin{tabular}{|c|c|c|}
\hline
\textbf{Training dataset} & \textbf{IoU} & \textbf{SKIoU} \\
\hline
\multicolumn{3}{|c|}{\textbf{Microtubules}} \\
\hline
Annotated Dataset \cite{liu2018densely} & 0.9439 & 0.9775 \\
Annotated + Synthetic dataset & 0.9527 & 0.9849\\
Annotated + Synthetic + Filament-aware Loss & \textbf{0.9542} & \textbf{0.9874}  \\
\hline
\multicolumn{3}{|c|}{\textbf{Actin Filaments}} \\
\hline
Original Dataset \cite{liu2018densely} & 0.9039 & 0.9575 \\
Annotated + Synthetic dataset & 0.9334 & 0.9631\\
Annotated + Synthetic + Filament-aware Loss & \textbf{0.9474} & \textbf{0.9658}  \\
\hline
\end{tabular}
\label{tab:results}
\end{center}
\end{table}

\section{Conclusion}
In this work, we propose a conditional generative framework to generate synthetic microscopy images of filamentous structures from binary masks. Our work addresses the data shortage problem of training segmentation models for filamentous structures in biological images. Our proposed model not only learns the imaging properties of filamentous structures in fluorescence microscopy but also learns the fluorescence background variations. Our qualitative results have shown that our generated images successfully simulate the mechanism of fluorescence microscopy imaging process. Our quantitative experiments further demonstrates that the inclusion of our synthetic images for training significantly improves the segmentation performance. In this work, we adopted a GAN-based framework instead of diffusion--based models for computational efficiency and faster training with limited data resources. In future work, we will explore diffusion-based generative architectures with few shot learning to further enhance the diversity and controllability of the generated images. We also aim to extend our framework to other thin-structure datasets, such as vascular, and to develop automated tools for quantitative analysis of filament morphology and dynamics. 

\bibliographystyle{unsrt}
\bibliography{egbib}

\end{document}